\documentclass[10pt,twocolumn,letterpaper]{article}

\usepackage{iccv}
\usepackage{times}
\usepackage{epsfig}
\usepackage{graphicx}
\usepackage{amsmath, bm}
\usepackage{amssymb}
\usepackage{booktabs,colortbl}

\usepackage[pagebackref=true,breaklinks=true,letterpaper=true,colorlinks,bookmarks=false]{hyperref}

\iccvfinalcopy % *** Uncomment this line for the final submission

\def\iccvPaperID{558} % *** Enter the ICCV Paper ID here

\ificcvfinal\fi
\begin{document}

%%%%%%%%% TITLE
\title{Fine-grained Recognition in the Wild:\\ A Multi-Task Domain Adaptation Approach}

\author{Timnit Gebru \quad Judy Hoffman \quad Li Fei-Fei\\
CS Department Stanford University\\
{\tt\small \{tgebru, jhoffman, feifeili\}@cs.stanford.edu}
}

\maketitle
%\thispagestyle{empty}
%%%%%%%%% ABSTRACT
\begin{abstract}
While fine-grained object recognition is an important problem in computer vision, current models are unlikely to accurately classify objects in the wild. These fully supervised models need additional annotated images to classify objects in every new scenario, a task that is infeasible. However, sources such as e-commerce websites and field guides provide annotated images for many classes. In this work, we study fine-grained domain adaptation as a step towards overcoming the dataset shift between easily acquired annotated images and the real world. Adaptation has not been studied in the fine-grained setting where annotations such as attributes could be used to increase performance. Our work uses an attribute based multi-task adaptation loss to increase accuracy from a baseline of $4.1\%$ to $19.1\%$ in the semi-supervised adaptation case. Prior domain adaptation works have been benchmarked on small datasets such as~\cite{office_dataset} with a total of $795$ images for some domains, or simplistic datasets such as~\cite{svhn} consisting of digits. We perform experiments on a subset of a new challenging fine-grained dataset consisting of $1,095,021$ images of $2,657$ car categories drawn from e-commerce websites and Google Street View.

\end{abstract}

%%%%%%%%% BODY TEXT
\section{Introduction} 
%\tg{Resubmission from CVPR. Blue shows things that have changed between the CVPR and ICCV versions or are in the process of changing.}}
The ultimate goal of image recognition is to recognize all objects in the world, as they appear in their natural environments. An even more difficult task, fine-grained recognition, aims to distinguish between objects in the same category (e.g. different bird species or car brands). Current state-of-the-art fine-grained classification methods~\cite{fg1,fg2,fg3,jon} focus on fully supervised learning regimes: a setting where human annotated images are available for all object categories of interest. To enable these methods, datasets have been proposed to train models recognizing all categories and scenes ~\cite{imagenet, krishnavisualgenome, places}, or focus on the fine-grained recognition task~\cite{cubs,dogs,comp_cars,flowers, stanford_cars}.

Models trained on these datasets are capable of outperforming humans when evaluated on benchmark tasks such as~\cite{imagenet, olga}. However, this evaluation paradigm ignores a key challenge towards the development of real world object classification models. Namely, fixed datasets such as ImageNet or Birds offer a sparse and biased sample of the world~\cite{dataset_bias}. Thus, to achieve comparable performance in real-world settings, fully supervised models trained with these datasets need additional annotated data from each new scenario. However, collecting images capturing all possible appearances of an object in a constantly changing real world environment is infeasible. The large number of possible images makes it prohibitively expensive to obtain labeled examples for every object category in the real world. Moreover, this annotation burden is amplified when we consider recognition for fine-grained categories. In this setting, only experts are able to provide our algorithms with labeled data.

\begin{figure}[t]
\includegraphics[width=\linewidth]{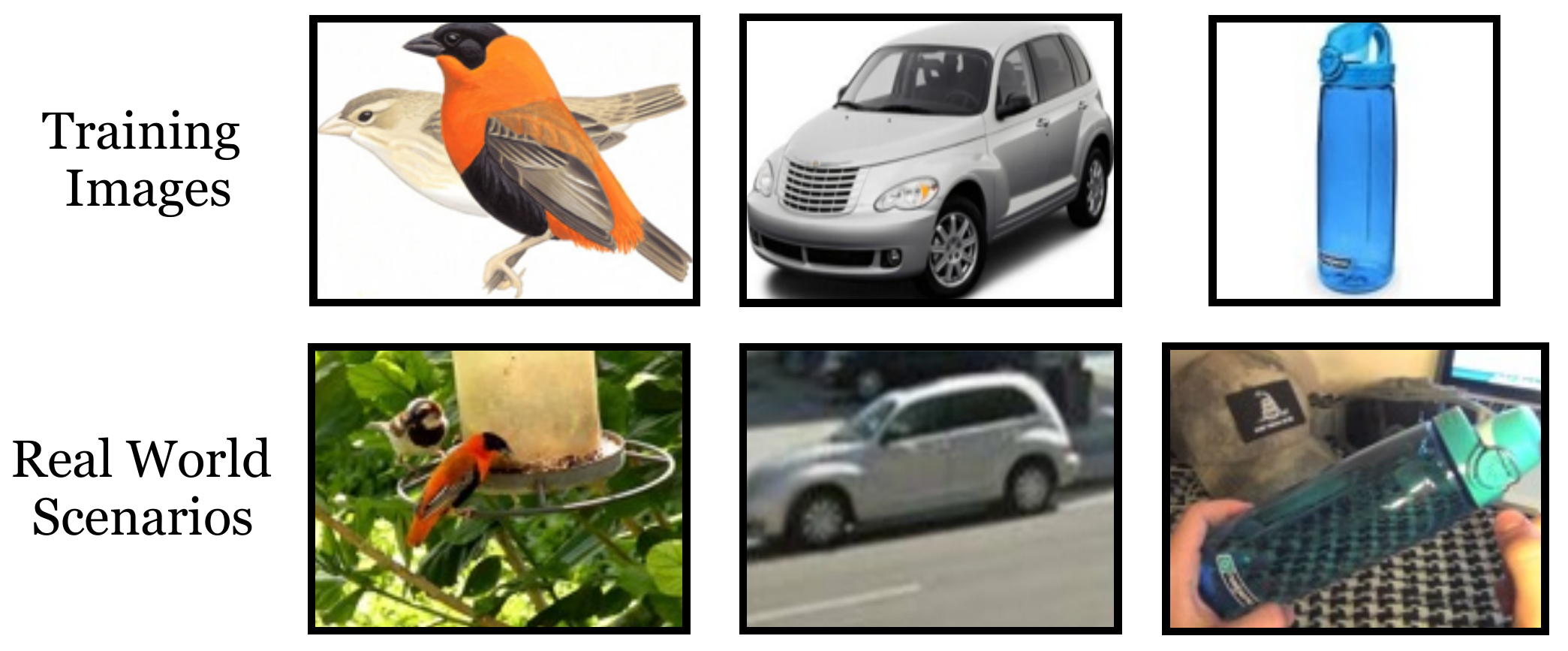}
\caption{
We aim to recognize fine-grained objects in the real world without requiring large amounts of expensive expert annotated images.
Instead, we propose training fine-grained models using cheaper annotated data such as field guides or e-commerce web sources (see \textit{top row}). We adapt the learned models to our task using only a sparse set of annotations in the real world.
} 
\label{fig:pull_figure}
\end{figure}

Fortunately, freely available sources of paired images and category labels exist for many objects we may want to recognize. For example, images and annotations from a field guide can be used to train a model recognizing various bird species in the wild (Fig.~\ref{fig:pull_figure} \textit{(top row)}). Similarly, annotated car images on e-commerce websites can be used to train a model distinguishing between different types of cars in unstructured urban environments (Fig.~\ref{fig:pull_figure} \textit{(middle row)}). However, images from these sources have different statistics from those we may encounter in the real world. And this statistical difference can cause significant degradation of model performance~\cite{dataset_bias, office_dataset, theory06}.

In this work, we study fine-grained domain adaptation as a step towards overcoming the dataset shift between easily acquired annotated images and the real world. To our knowledge, adaptation has not been studied in the fine-grained setting where it is especially expensive to obtain image annotations. In this scenario, many of our categories may be related to one another in some known hierarchical way. For example, multiple distinct car varieties may share the same body type or the same make.

Our contributions are two fold: first, we propose a new multi-task adaptation approach which explicitly benefits from these known cross-category relationships. Our model consists of a multi-task adaptation objective which simultaneously learns and adapts recognition at the attribute and category level. We first show that our objective effectively regularizes the source training and hence improves the generalization of the source model to the target domain. Then, for the task of semi-supervised adaptation (i.e. when category labels are only available from a subset of the classes in the target domain), we exploit the fact that labels will often exist for all attributes. For example, while annotated target images for a 1998 Honda Accord sedan may not be available, some images of other Hondas and sedans are likely in our dataset. In this way, we are able to apply different adaptation techniques at the class and attribute levels. 

Our second contribution characterizes a large scale fine-grained car dataset for domain adaptation. While this dataset was introduced by~\cite{aaai} in the context of fine-grained detection, it has not been used in adaptation.  We perform experiments on a subset of $170$ out of $2,657$ classes (a total of $71,030$ images) and show significantly improved performance using our method. While visual domain adaptation has been well studied~\cite{office_dataset,aytar-iccv11, gong-cvpr12, tzeng2015iccv, ganin2015}, most approaches focus on adapting between relatively small data sources consisting of tens of object categories and hundreds of images in total~\cite{office_dataset, crossdataset, svhn}. The use of such small datasets in developing adaptation algorithms makes it difficult to reliably benchmark these algorithms. To our knowledge, our work is the first to study this important problem on a large scale, real-world dataset and in the fine-grained scenario.

\section{Related Work}
\textbf{Fine-Grained object recognition.} While fine-grained image recognition is a well studied problem~\cite{fg1,fg2,fg3,fg4,fg5,fg6,fg7,fg8,fg9,fg10}, its real world applicability is hampered by limited available data. Works such as~\cite{jon} have used large-scale noisy data to train state-of-the-art fine-grained recognition models. However, these models are unlikely to generalize to real world photos because they are trained with images derived from field guides or product shots. Similarly, standard fine-grained datasets such as~\cite{cubs} and~\cite{birdsnap} are derived from a single domain. Due to the large
variation in object appearance between the real world and these datasets, models trained on these images are unlikely to generalize well to real world objects. 

\textbf{Domain adaptation.} Domain adaptation works enhance the performance of models trained on one domain (such as product shot images) and applied to a different domain (such as real world photos). Since the theoretical framework provided by~\cite{theory06}, many computer vision works have published algorithms for unsupervised domain adaptation: i.e. a task where no labeled target images are available during training~\cite{pixel,residual,backprop,lightweight,separation, deep_dom}. Most methods strive to learn a classifier with domain invariant features~\cite{tzeng2015iccv,adversarial, mjordan} . Long et al. relax the assumption of a single classifier for both source and target images and instead use $2$ classifiers with a residual connection~\cite{residual}. 
While these works focus on unsupervised domain adaptation,~\cite{tzeng2015iccv} performs semi-supervised adaptation, transferring knowledge from classes with labeled target images to those without. To our knowledge, there have been no studies of visual adaptation in the fine-grained setting. Our work builds on~\cite{tzeng2015iccv}'s method to show that attribute level softlabel transfer and domain confusion significantly boost performance in this scenario. 

\textbf{Attributes, structured data and multitask learning.} Attributes have been used to improve object classification in~\cite{semantic_att} and perform zero shot learning in~\cite{zero_shot,att5}. Kodirov et al.~\cite{zsl_adaptation} uses sparse coding and subspace alignment techniques to perform zero shot learning when images are sourced from multiple domains. We draw inspiration from these works and leverage attributes to improve performance in unsupervised and semi-supervised domain adaptation. In contrast to~\cite{att_da}'s adaptation of user specified attributes, we use labels shared between different fine-grained categories to facilitate class level transfer. While prior works such as~\cite{att2,att3,att4} focus on attribute learning, our goal is to improve adaptation using ground truth attribute labels. 

Our method to enforce consistency between attribute and class predictions is similar in spirit to a number of works exploiting label structure~\cite{jia,structure2}.~\cite{jia} uses Hierarchy and Exclusion (HEX) graphs to encapsulate semantic relations between pairs of labels. We use a KL divergence loss between predicted label distributions instead of hard constraints. 

Finally, some prior works have shown that learning multiple tasks can improve generalization for each task. For example,~\cite{segmentation} found that a multi-task network for segmentation improves object detection results as a bi-product. Similarly,~\cite{translation} showed that a machine learning to translate multiple languages performs better on each language. We observe similar results where a multi-task adaptation approach using attributes improves class level performance.

%\vspace{-0.2cm}
%\textbf{Adaptation datasets.} Although domain adaptation has been studied extensively in the natural language processing community~\cite{frustrating,hierarchical}, the first computer vision dataset for this task was introduced in 2010~\cite{office_dataset}. This dataset consists of a total of $4,652$ images of $31$ types of objects found in the office (such as bags, laptops and screens). And it is still the standard domain adaptation dataset used today. A slightly larger dataset was more recently released with cross dataset category overlaps~\cite{crossdataset}, but it still only contains $40$ basic level categories. CompCars, a large scale dataset of $214,345$ images and $1,687$ classes of cars from $2$ sources (e-commerce websites and surveillance cameras), can potentially be used for domain adaptation~\cite{cars}. However, the images from surveillance cameras only have front and rear views of the cars. Furthermore, images are annotated at the car model level. I.e., visually distinct vehicles from different years such as a 1990 Honda Accord and a 2015 Honda Accord are grouped into the same class. This makes the dataset less suitable for fine-grained  classification.
\begin{figure*}
    \centering
      \includegraphics[width=\linewidth]{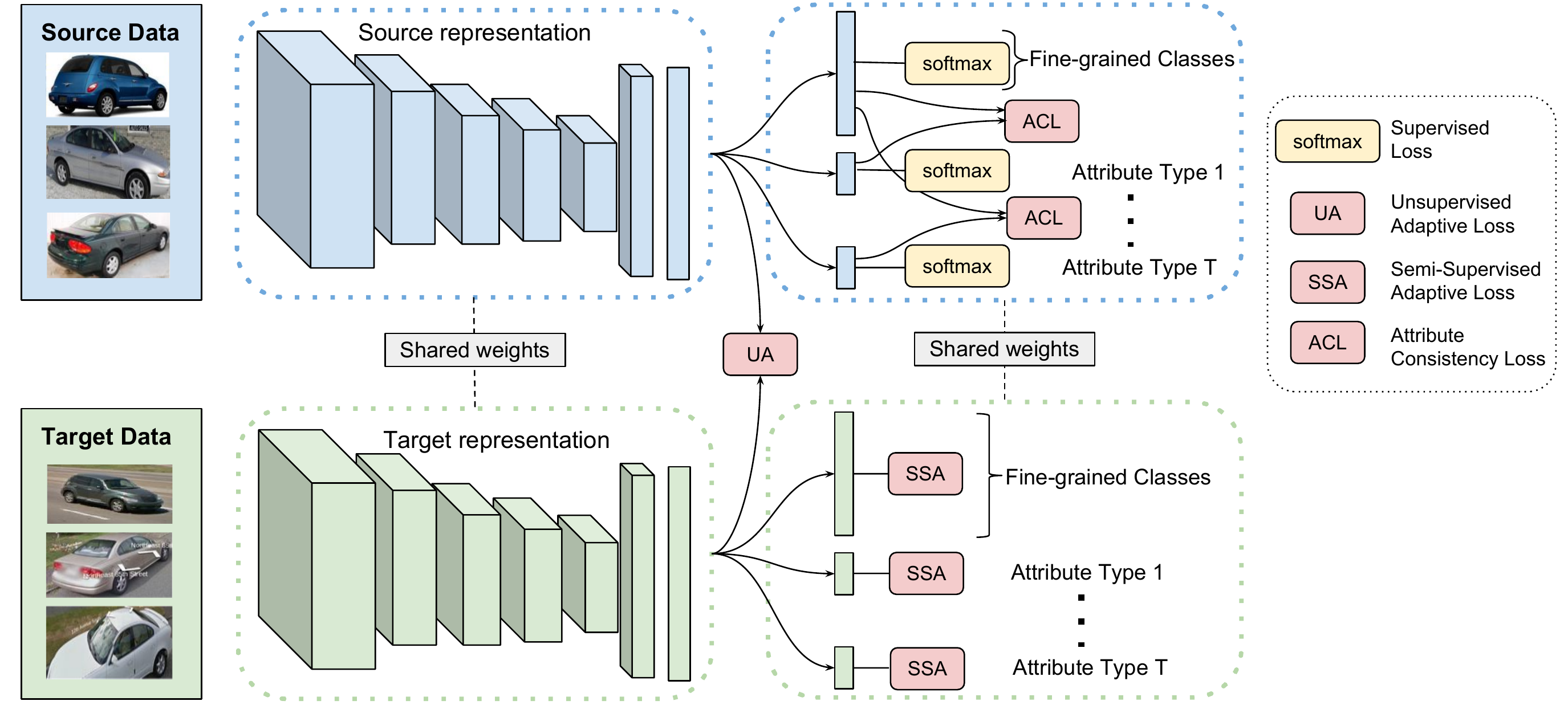}
    \caption{ Our architecture for unsupervised and semi-supervised domain adaptation. Two CNNs based on ~\cite{alexnet} with shared weights classify source and target images. The fc7 feature maps of labeled source and target images are input into independent softmax classifiers classifying each attribute and fine-grained class of the image. Any unsupervised adaptive loss such as domain confusion (denoted as UA)~\cite{tzeng2015iccv} can be used to further improve adaptation. When labeled target images are available, semi-supervised adaptive loss (denoted as SSA) such as the soft label loss of~\cite{tzeng2015iccv} can be performed at the attribute, as well as fine-grained level. An attribute consistency loss (denoted as ACL) encourages the fine-grained and attribute classifiers to predict consistent labels.}
    \label{fig:architecture}
\end{figure*} 
%\section{Large-Scale Car Dataset for Fine-Grained Domain Adaptation}
%\label{sec:dataset}
%\input{dataset}

\section{Multi-Task Domain Adaptation for Fine-Grained Recognition}
In the fine-grained classification setting, obtaining labels for every single class is infeasible. However, classes often share attributes. For instance, a Beagle and a Jack Russell terrier are both small dogs while a Bearded Collie and Afghan Hound are both shaggy dogs. In the general object classification setting, a taxonomic tree such as WordNet can be used to group categories and obtain labels at multiple levels in the hierarchy. Thus, while the target domain may not have labels for every leaf node class, we are more likely to have images annotated at higher levels in the hierarchy.

We leverage these additional annotations in a multi-task objective, providing regularization and additional supervision. 
Specifically, we minimize a multi-task objective consisting of softmax classification losses at the fine-grained and attribute level. In our architecture shown in Fig.~\ref{fig:architecture}, this is achieved by having multiple independent softmax layers that perform attribute level, in addition to category level, classification. We add an attribute consistency loss to prevent the independent classifiers from predicting conflicting labels. Any unsupervised adaptive loss (denoted as UA) in Fig.~\ref{fig:architecture} can be used in conjunction with our method. Similarly, when target labels are available for some classes, any semi-supervised adaptive loss (denoted as SSA) can be added at the class and attribute levels. Here, we apply our method to~\cite{tzeng2015iccv} to evaluate its efficacy.

\subsection{CNN Architecture for Multi-Task Domain Transfer}
\label{sec:multitask}
We give an overview of our architecture for semi-supervised domain adaptation shown in Fig.~\ref{fig:architecture}. Our model is trained using annotated source images for all classes, which we denote as $\{x_S,y_S\}$, and labeled and unlabeled target images, $\{x_T,y_T\}$. $x_S$, $x_T$ are source and target image samples respectively and $y_S$, $y_T$ are their associated labels. Our goal is to train a model classifying images \{$x_T$\} for fine-grained categories with no labeled target images. We denote the number of target images as $N_T$ and the number of labeled target images as $N_{TL}$. $N_{TL}=0$ and $N_{TL}=N_T$ in the unsupervised and fully supervised adaptation settings respectively.
%In the unsupervised adaptation setting $N_{TL}=0$ and in the supervised adaptation setting $N_{TL}=N_T$. 
Only a subset of the target images are labeled in the semi-supervised adaptation setting resulting in $N_{TL} < N_T$. 

In addition to class labels $y_S$, $y_T$, we also have attribute level annotations $y_{Sa}$, $y_{Ta}$ for source and target images respectively. There are at least as many labeled source and target images available for each attribute $a$, as each class $c$. This implies that even when no labeled target images are available for class $c$, there are labels for classes with similar attributes to $c$. 
%Thus, especially in the unsupervised and semi-supervised domain adaptation settings, we expect fine-tuning using attribute labels to provide a significant performance gain. With this intuition, 
We optimize a multi-task loss with $3$ components: a softmax classification loss at the fine-grained and attribute levels, an attribute consistency loss, and any unsupervised or semisupervised adaptation loss.
%a global domain confusion loss, and a softlabel transfer loss performed at the class and attribute level. 

\subsection{Classification Loss}
\label{sec:classification_loss}
We start with a CNN following the architecture of~\cite{alexnet}, taking $\{x_S,y_S\}$ and $\{x_T,y_T\}$ as inputs. We denote the parameters of this classifier as $\theta_{rep}$. 
Let each attribute $a$ have $a_K$ categories. We have $N_a$ attribute classifiers $f_{a_n}$ parametrized by $\theta{a_n}, n=1...N_a$. These classifiers operate on the image feature map $f (x,y;\theta_{rep})$ produced by our CNN. $N_a$ is the number of attributes and $x,y$ are an input image and its associated label respectively. We minimize $N_a$ softmax losses:
\begin{equation}
    L_{a_n}(x,y;\theta_{rep},\theta_{a_n})=-\sum_{a_k=1}^{a_K}{\bm{1}[y_a=a_k]\log p_{ak}}
\end{equation}
\setlength\belowdisplayskip{4pt}
where $y_a$ is the ground truth label for image $x$ and attribute $a$, and $p_a=[p_{a1},...p_{aK}]$ is the softmax of the activations of attribute classifier $f_{a_n}$. I.e., $p=softmax(\theta^{T}_{a_n}f(x;\theta_{rep}))$.

In addition to attribute level softmax losses, we minimize a softmax classification loss at the fine-grained level.
With $K$ classes, and a fine-grained classifier parametrized by $\theta_{C}$ operating on feature map $f(x,y; \theta_{rep})$, we minimize the loss:
\begin{equation}
    L_{C}(x,y;\theta_{rep},\theta_{C})=-\sum_{k=1}^{K}{\bm{1}[y=k]\log p_{k}}
\end{equation}

Our final multi-task softmax loss is the weighted sum of the attribute and fine-grained softmax losses.  Omitting parameters for simplicity of notation, 

\begin{equation}
L_{softmax} =\sum_{n=1}^{N_a}{\alpha_n L_{a_n}} + \alpha_c L_C
\end{equation}

\begin{figure}
\includegraphics[width=1\linewidth]%{figs/attribute_consistency.pdf}
{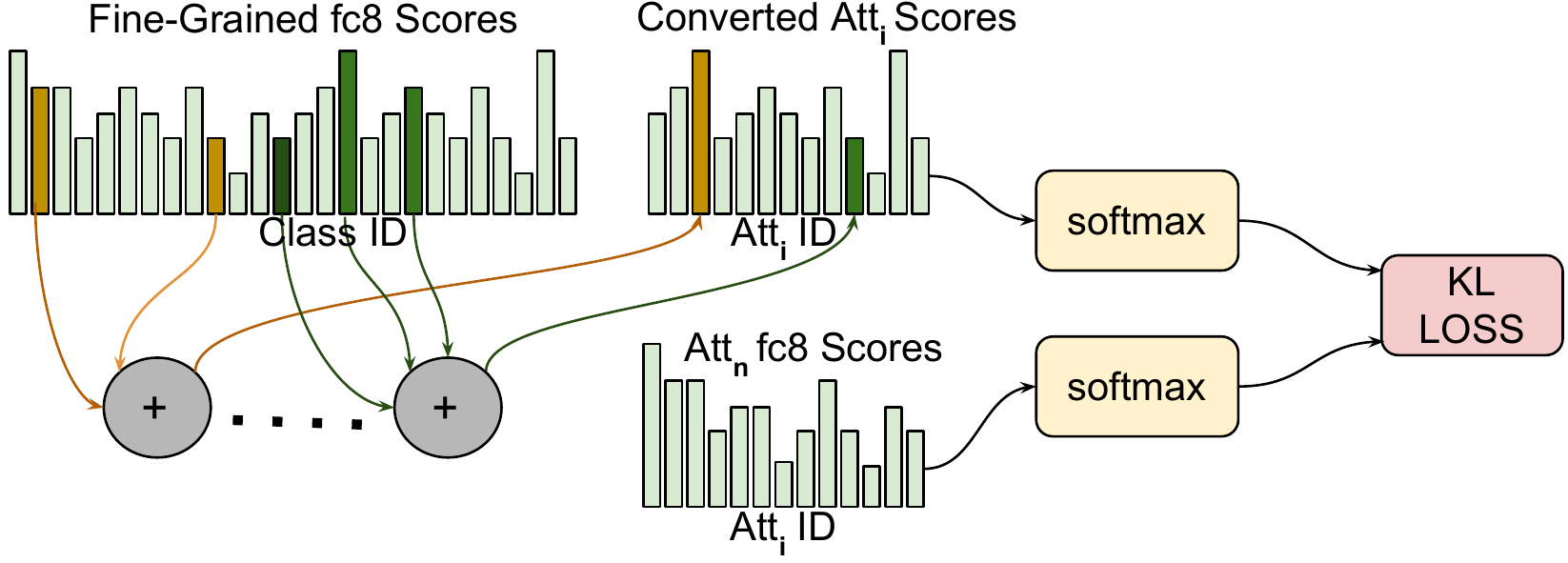}
\caption{An attribute consistency loss between the fine-grained and attribute classifiers encourages them to predict consistent results. For each of the $i$ attributes $Att_i$, $fc8$ scores from the fine-grained classifier are converted to scores across attributes. We minimize a KL divergence loss between the softmax of these attribute scores and the softmax output of the attribute classifier $fc8_{Att_i}$.}
\label{fig:attribute-consistency}
\end{figure}

\subsection{Attribute Consistency Loss}
\label{sec:consistency}
While our attribute and class classifiers are independently trained using ground truth labels, our pipeline so far poses no restrictions on how these classifications are related to each other. That is, the fine-grained classifier can output a class whose attributes are different from ones predicted by the attribute classifiers. However, we know that the attributes of the fine-grained class should be the same as those predicted by the independent attribute classifiers. To enforce this structure, we add an attribute consistency loss that penalizes differences between attributes predicted by the fine-grained and attribute classifiers. We minimize a symmetric version of the KL divergence between the distribution of attributes predicted by attribute classifier $a_n$ and those inferred by the fine-grained class classifier. Our procedure is visualized in Fig.~\ref{fig:attribute-consistency}. For each attribute $a$, we first convert scores across classes (fc8 output in~\cite{alexnet}) to ones across categories for that attribute. Let $f=[f_1,...f_k]$ consist of scores for $k$ classes. To obtain scores across attribute categories, we average values in $f$ belonging to classes from the same attribute. Averaging, rather than simply adding, scores mitigates the effect of dataset bias where some classes and attributes appear more frequently. We then compute a softmax distribution across attribute categories for attribute $a$,  $\hat{p}_a=[\hat{p}_{a1},...,\hat{p}_{aK}]$ using the computed attribute scores.

We define a consistency loss for each attribute $a$ as the symmetric version of the KL divergence between $\hat{p}_a$ and $p_a$:
\begin{equation}
L_{con_{a_n}}(x,\theta_{rep},\theta_{a_n}, \theta_{c})=\frac{1}{2}D_{KL}(p_a||\hat{p}_a)+\frac{1}{2}D_{KL}(\hat{p}_a||p_a)
\end{equation}

%\begin{equation}
%L_{con_{a_n}}(x,\theta_{rep},\theta_{a_n}, \theta_{c})=-\sum_{a_k=1}^{a_K}%{\bm{1}[a(c)=a_k]\log p_{ak}}
%\end{equation}
\begin{equation}
D_{KL}(p_a||\hat{p}_a)=\sum_{a_k=1}^{a_K}{p_{ak}\log \frac{p_{ak}}{\hat{p}_{ak}}}
\end{equation}
where attribute $a$ has $a_K$ categories as defined in~\ref{sec:classification_loss}. Since we are not trying to match a reference distribution and are only minimizing the distance between two distributions, we use a symmetric version of the KL divergence in our loss instead of cross-entropy loss. Omitting parameters for simplicity, the final consistency loss $L_{consistency}$ is a weighted sum of the losses for each attribute:

\begin{equation}
L_{consistency}=\sum_{n=1}^{N_a}{\beta_{a_n}L_{con_{a_n}}} 
\end{equation}

\subsection{Augmenting Existing Adaptation Algorithms with Attribute Loss}
\label{dc}
We can augment any existing adaptation algorithm with our attribute based losses to perform adaptation at the attribute as well as the class level. Here, we describe how we apply our method to~\cite{tzeng2015iccv}. To use our method with~\cite{tzeng2015iccv}, we add the domain confusion and softlabel losses introduced in~\cite{tzeng2015iccv}. The softlabel loss is only used in the semi-supervised setting where labeled target images are available for some classes. However, in addition to a softlabel loss $L_{csoft}$ at the fine-grained level, we also minimize the softlabel objective $L_{asoft}$ for each attribute $a$. This allows us to leverage attribute level annotations that exist for classes with no labeled target images. Denoting the domain confusion loss as $L_{conf}$, our final objective is a weighted sum of $L_{csoft}$, $L_{asoft}$, $L_{conf}$, $L_{softmax}$ and $L_{consistency}$.

\section{Evaluation}
\begin{figure}
\includegraphics[width=\linewidth]{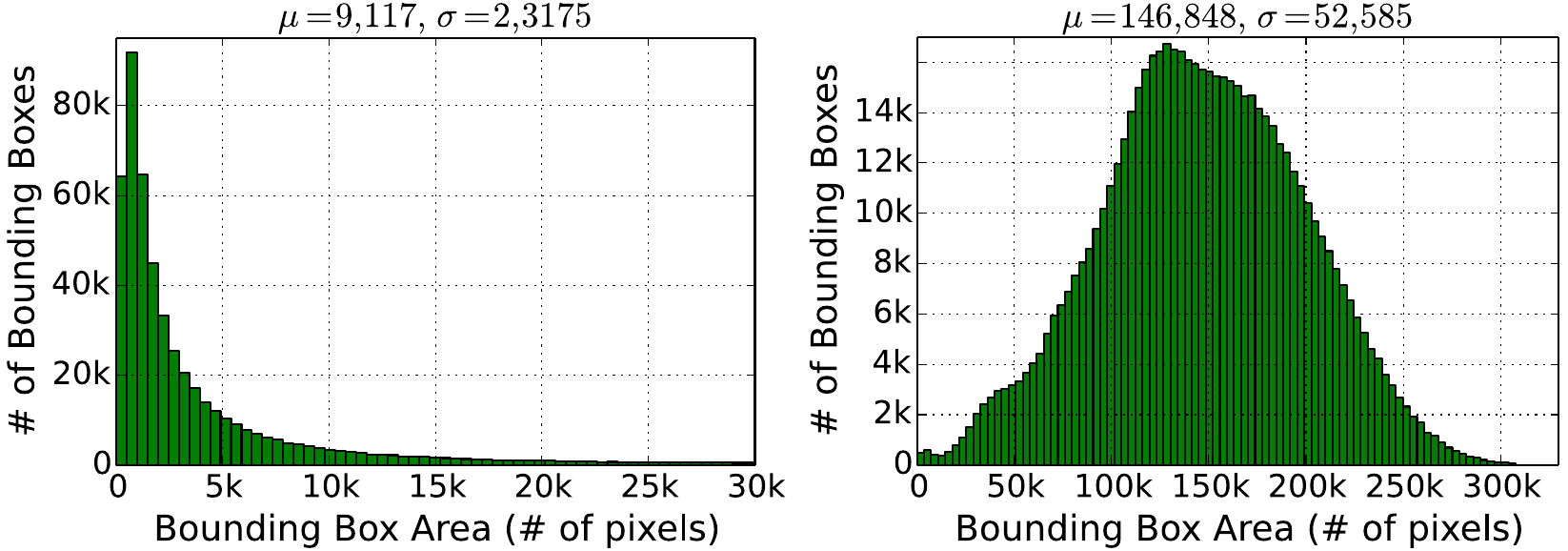}
\caption{Histogram of \textit{GSV} (left) and \textit{web} (right) bounding box sizes. While cars in \textit{GSV} images are typically small (with an average size of $9,117$ pixels), those in \textit{web} images are much larger, occupying an average of $146,848$ pixels.}
\label{fig:dataset_bboxes}
\end{figure}

\begin{figure}
\includegraphics[width=\linewidth]{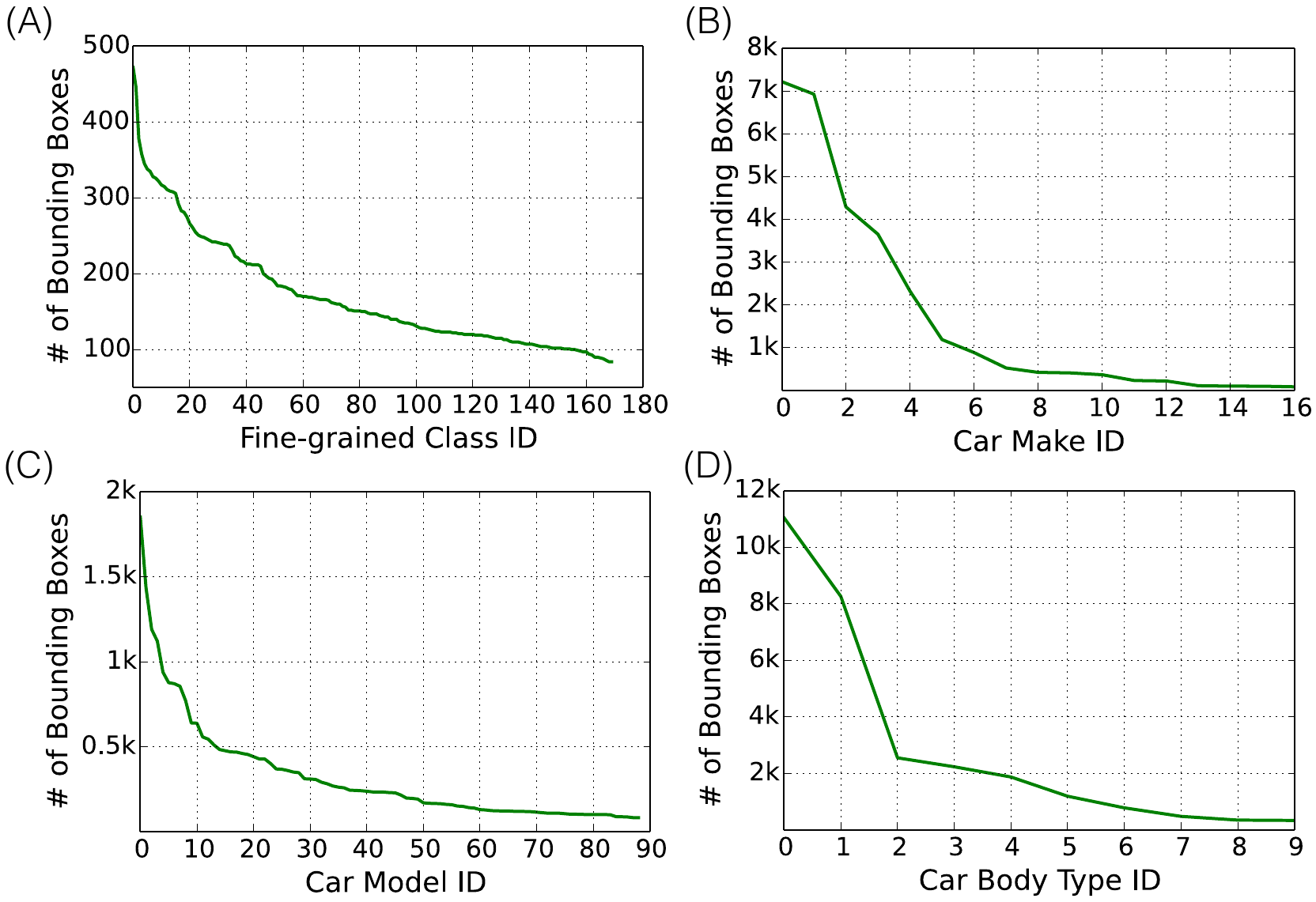}
\caption{The distribution of \textit{GSV} images for each class (A), each make (B), each model (C) and each body type (D) for the subset of the car dataset used in our evaluation. While each fine-grained class has less than $500$ labeled images, some body types have close to $12,000$ labeled \textit{GSV} images (D).}
\label{fig:label_dist}
\end{figure}

\begin{figure*}[ht!]
\includegraphics[width=\linewidth]{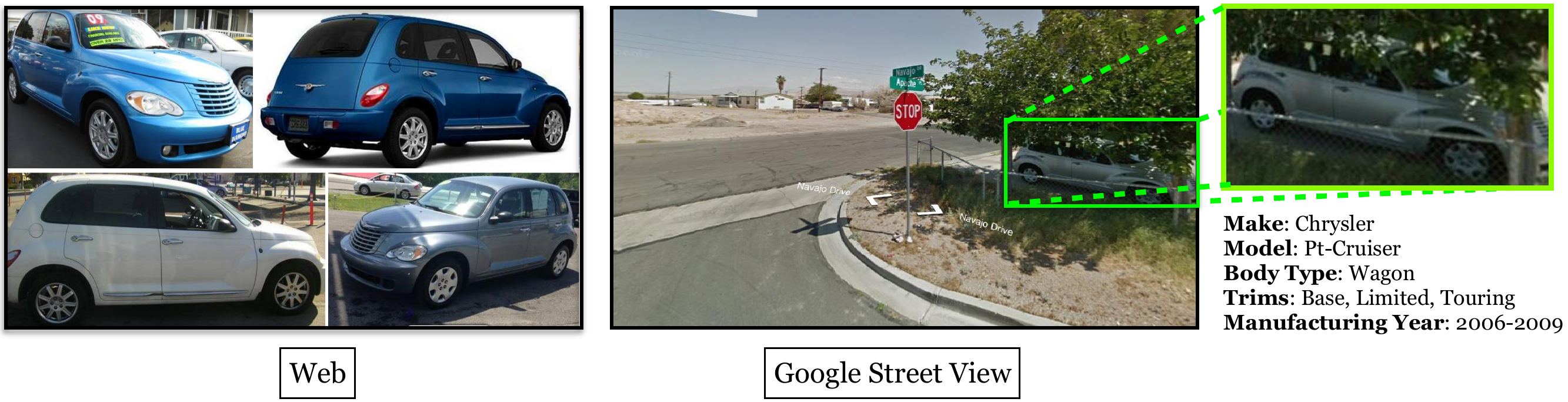}
\caption{Examples of \textit{web} and \textit{GSV} images for one type of car in our dataset. \textit{Web} images are typically un-occluded with a high resolution while \textit{GSV} images are blurry and occluded.}
\label{fig:dataset_sources}
\end{figure*}
We evaluate our multi-task adaptation algorithm on two datasets, a recently proposed large scale car dataset~\cite{aaai} and the office dataset~\cite{office_dataset} augmented with attributes taken from the WordNet~\cite{wordnet} hierarchy.  
To test the efficacy of our attribute level adaptation approach, we modify an existing domain adaptation method, DC~\cite{tzeng2015iccv}, by adding our attribute level losses.

We use Caffe~\cite{caffe} in all of our experiments. Our source only models are initialized with ImageNet weights using the released CaffeNet model~\cite{caffe}. For experiments on the car dataset, we use equal weights across all our losses and a temperature of $2$ while calculating softlabel losses. We set the learning rate to $0.0001$ for all experiments and will release our custom layers for optimizing $KL$ divergence loss. For experiments on the office dataset, we set all loss weights to $1$ except for domain confusion loss whose weight was set to $0.1$.

\begin{table}
\begin{center}
\begin{tabular}{lccccc}
\toprule
\multicolumn{2}{c}{} & \multicolumn{4}{c}{\textbf{Accuracy (\%)}}\\
\cline{3-6}
\textbf{Train} & \textbf{Test} & \textbf{Class} & \textbf{Make} & \textbf{Model} & \textbf{Body}\\
\midrule \midrule
S & S  & $73.9$ & $85.0$ & $82.2$ & $92.0$\\ 
S & T  & $8.5$  & $36.2$ & $18.2$ & $59.7$\\
T & T  & $18.9$ & $51.9$ & $31.6$ & $73.9$\\ 
%Target & Source  & $5.5$  & $41.4$ & $19.0$ & $60.3$ \\
%S + T  & S & $75.5$ & $89.4$ & $81.7$ & $89.5$\\
S+T  & T & $27.9$ & $56.4$ & $41.1$ & $75.8$\\
\bottomrule
\end{tabular}
\end{center}
\caption{We quantify the amount of domain shift between the \textit{web} source domain (S) and \textit{GSV} target domain (T).
Training on source and evaluating on target shows a significant performance drop.
Accuracies are shown for models trained at the fine-grained class, make, model and body-type level. There are $170$ fine-grained classes, $89$ models, $17$ makes and $10$ body-types in our dataset.}
\label{table:domain_shift}
\end{table}
\begin{table}
\begin{center}
\resizebox{\linewidth}{!}{%
\begin{tabular}{lcccc}
\toprule
\textbf{Model} & \textbf{Adapt} & \textbf{Attr} & \textbf{Consist} & \textbf{Acc} (\%)\\
\midrule
{Source CNN}  &  & & & $9.28$ \\
{Source CNN w/att} & & \checkmark & & 10.80\\
{Source CNN w/att+ACL}	& 	& \checkmark & \checkmark & ${14.37}$\\ 
{DC~\cite{tzeng2015iccv}} & \checkmark &  & & $14.98$ \\
{DC~\cite{tzeng2015iccv} w/att+ACL} & \checkmark & \checkmark & \checkmark & $\textbf{19.05}$\\
\bottomrule
\end{tabular}%
}
\end{center}
\caption{\textbf{Cars$\rightarrow$GSV Unsupervised Adaptation:} We report multi-class accuracy for all classes in the GSV validation set and demonstrate the effectiveness of incorporating our attributes and consistency loss into the baseline and adaptive methods.}
\label{table:cars_unsup}
\end{table}
\begin{table}
\begin{center}
\resizebox{\linewidth}{!}{%
\begin{tabular}{lcccc}
\toprule
\textbf{Model} & \textbf{Adapt} & \textbf{Attr} & \textbf{Consist} & \textbf{Acc} (\%)\\
\midrule
{S+T CNN}  &  & & & 4.12 \\
%{S+T CNN w/att} & & \checkmark & &  \\
{S+T CNN w/att+ACL}	& 	& \checkmark & \checkmark & 7.45\\ 
{DC~\cite{tzeng2015iccv}} & \checkmark &  & &  12.34\\
{DC~\cite{tzeng2015iccv} w/att+ACL} & \checkmark & \checkmark & \checkmark & $\textbf{19.11}$\\
\bottomrule
\end{tabular}%
}
\end{center}
\caption{\textbf{Cars$\rightarrow$GSV Semi-supervised Adaptation:} We report multi-class accuracy for the held-out unlabeled classes in the GSV validation set and demonstrate the effectiveness of incorporating our attributes and consistency loss into the baseline and adaptive methods.}
\label{table:cars_semi}
\end{table}
\begin{table}
\begin{center}
\resizebox{\linewidth}{!}{%
\begin{tabular}{lcccc}
\toprule
\textbf{Model} & \textbf{Adapt} & \textbf{Attr} & \textbf{Consist} & \textbf{Acc} (\%)\\
\midrule
{Source CNN}  &  & & &  60.9\\
{Source CNN w/att} & & \checkmark & &  59.5\\
{Source CNN w/att+ACL}	& 	& \checkmark & \checkmark & 61.2\\ 
{DC~\cite{tzeng2015iccv}} & \checkmark &  & &  61.1\\
{DC~\cite{tzeng2015iccv} w/att+ACL} & \checkmark & \checkmark & \checkmark & $\textbf{62.4}$\\
\bottomrule
\end{tabular}%
}
\end{center}
\caption{\textbf{Amazon$\rightarrow$Webcam Unsupervised Adaptation:} We report multi-class accuracy for the full Webcam dataset and demonstrate the effectiveness of incorporating our attributes and consistency loss into the baseline and adaptive methods.}
\label{table:office_unsup}
\end{table}

\begin{table}
\begin{center}
\resizebox{\linewidth}{!}{%
\begin{tabular}{lcccc}
\toprule
\textbf{Model} & \textbf{Adapt} & \textbf{Attr} & \textbf{Consist} & \textbf{Acc} (\%)\\
\midrule
{S+T CNN}  &  & & &  {45.5}\\
%{S+T CNN w/att} & & \checkmark & &  \\
{S+T CNN w/att+ACL}	& 	& \checkmark & \checkmark & 45.3\\ 
{DC~\cite{tzeng2015iccv}} & \checkmark &  & &  47.0\\
{DC~\cite{tzeng2015iccv} w/att+ACL} & \checkmark & \checkmark & \checkmark & $\textbf{51.8}$\\
\bottomrule
\end{tabular}%
}
\end{center}
\caption{\textbf{Amazon$\rightarrow$Webcam Semi-supervised Adaptation:} We report multi-class accuracy for the held-out unlabeled classes in the Webcam dataset and demonstrate the effectiveness of incorporating our attributes and consistency loss into the baseline and adaptive methods.}
\label{table:office_semi}
\end{table}

\subsection{Large scale car dataset}
The car dataset introduced in~\cite{aaai} consists of $1,095,021$ images of $2,657$ categories of cars from $4$ sources: craigslist.com, cars.com, edmunds.com and Google Street View. We refer to images from craigslist.com, cars.com and edmunds.com as \textit{web} images and those from Google Street View as \textit{GSV} images. As shown in Fig.~\ref{fig:dataset_sources}, cars in \textit{web} images are large and typically un-occluded whereas those in \textit{GSV} are small, blurry and occluded. The difference in image size is apparent in Fig.~\ref{fig:dataset_bboxes} which shows a histogram of bounding box sizes in \textit{GSV} and \textit{web} images. These large variations in pose, viewpoint, occlusion and resolution make this dataset ideal for a study of domain adaptation, especially in the fine-grained setting. 
In addition to the category labels, each class is accompanied by metadata such as the make, model body type, and manufacturing country of the car. 

\subsection{Quantifying Domain Shift on the Car Dataset}
In any adaptation experiment, it is crucial to first understand the nature of the discrepancy between the different sources of data. Following the standard set by~\cite{office_dataset}, we quantify this shift in the car dataset by training a sequence of models and evaluating both within and across domains. We perform all of our experiments on a subset consisting of $170$ of the most common classes in the dataset, particularly those with at least $100$ target images per class. This ensures that we have enough images to reliably evaluate our algorithm. 

In particular, we train a source only model and find that while accuracy is relatively high when evaluating within the source \textit{web} domain ($73.9\%$), performance catastrophically drops when evaluating within the \textit{GSV} target domain. To aid in analyzing our semi-supervised domain adaptation experiments, we train a target only model using all available \textit{GSV} labels. This model serves as an oracle for our adaptation experiments which use a reduced set of labeled images. As shown in Tab.~\ref{table:domain_shift}, the target only model significantly outperforms our source only model indicating a large shift between the two domains. Finally, we train a joint fully supervised source and target model to test whether \textit{web} and \textit{GSV} data are complementary. Indeed, the joint model outperforms even the fully supervised target only model. This indicates that annotated images from the source domain will be a useful resource to train models classifying target images. Thus, in the next set of experiments, we evaluate our adaptation solutions.

\begin{figure}
\includegraphics[width=\linewidth]{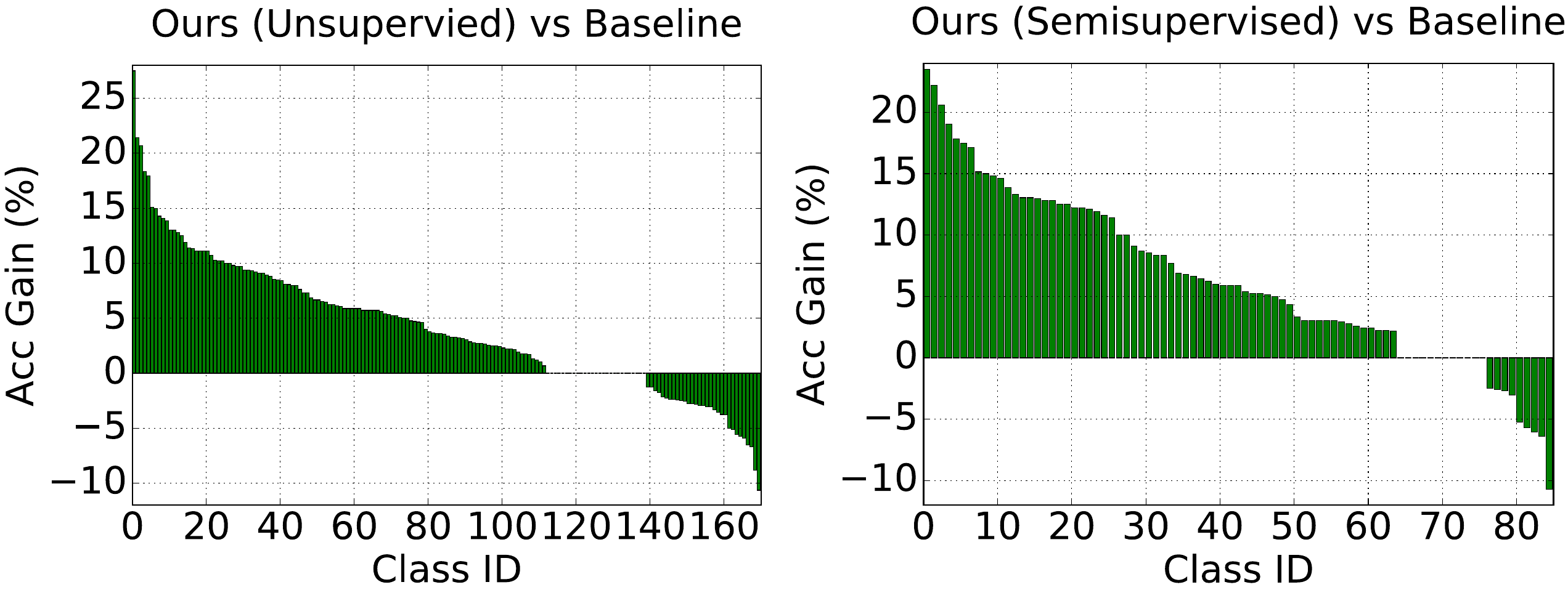}
\caption{The difference in accuracy per class between our models and baselines on the car dataset. $66\%$ of all fine-grained categories see a gain in accuracy in the unsupervised setting (\textit{left}). Similarly, in the semi-supervised setting, our model improves classification accuracy on $75\%$ of the held-out classes (\textit{right}).}
\label{fig:acc_gain} 
\end{figure}

\begin{figure}
\includegraphics[width=\linewidth]{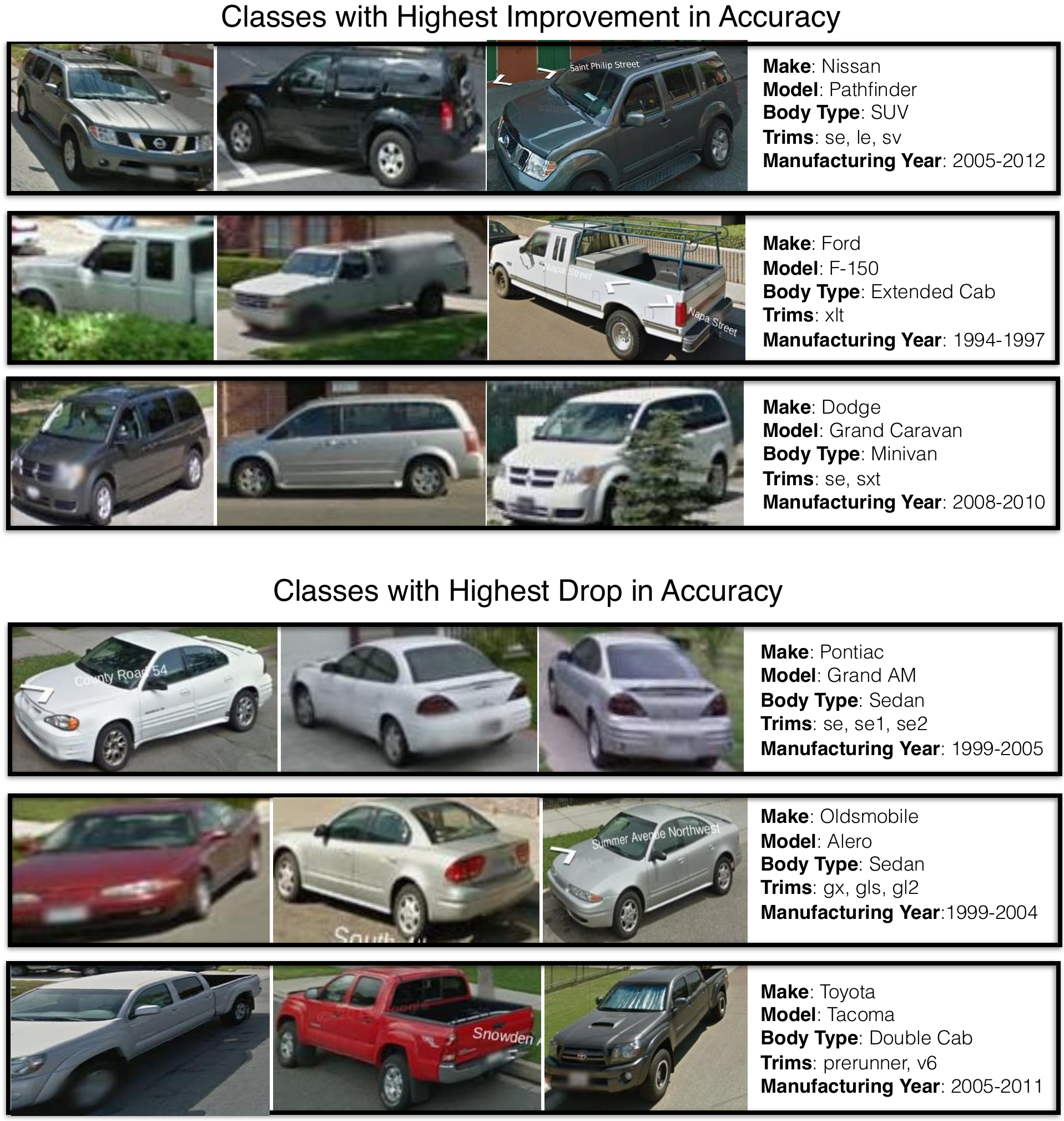}
\caption{Example images for classes resulting in the highest accuracy gain with our method (\textit{top}), and the highest accuracy drop with our method (\textit{bottom}) on the car dataset. The class with the highest accuracy gain is the 2008-2010 Dodge Grand Caravan while the 2005-2011 Toyota Tacoma sees the highest accuracy loss.}

\label{fig:results_fig}
\end{figure}
For each of these experiments, we train models using fine-grained class as well as make, model and body type labels.  There are $10$ body types, $17$ makes and $89$ models in the subset of the dataset used for our experiments. 

\subsection{Multi-Task Adaptation on the Car Dataset}
A real world domain adaptation pipeline should leverage the availability of labeled target images for popular fine-grained objects, to improve classification performance on classes whose labels are difficult to obtain. With this motivation, we partition the target data into labeled and unlabeled sets to perform semi-supervised domain adaptation experiments. We first sort the fine-grained classes by the number of target images they have. We then use images for the top $50\%$ of target classes ($85$ classes) with the highest number of labels in conjunction with source images for all classes as labeled training data. Our test data comprises of images for the $50\%$ of classes with the least number of labels. Thus, no labeled target images from the held-out classes are used in training the models used in semi-supervised adaptation experiments.

Table~\ref{table:cars_unsup} shows classification accuracies for various baseline methods as well as our architecture. Our baselines are source only and DC~\cite{tzeng2015iccv} adaptive models. We also compare our full model to one without attribute consistency loss. In all cases, our attribute level adaptation mechanism drastically improves performance. For example, in the unsupervised adaptation scenario, we see a $\sim10\%$ gain. To ensure that our attribute loss indeed aids adaptation and does not solely improve the baseline classifier, we also train a CNN that solely incorporates non-adaptation based components of our loss: i.e, $L_{softmax}$ and $L_{consistency}$. While attributes indeed improve the baseline model (accuracy jumps from $9.28\%$ to $14.37\%$), they also improve adaptation. For example, domain confusion increases accuracy by $\sim5\%$ without attributes but this improvement jumps to $\sim10\%$ with attributes. 

We see similar gains with our method in the semi-supervised adaptation setting. Training with a labeled subset of \textit{GSV} classes in addition to \textit{web} images generally reduces performance on the held-out \textit{GSV} classes; the model overfits to the labeled \textit{GSV} classes and becomes less generalizable. While domain confusion and softlabel loss combat this problem, we see the most significant improvement when these methods are used in conjunction with attribute level transfer: accuracy increases from $12.34\%$ to $19.11\%$. This confirms our intuition that using attribute labels helps our classifier learn domain invariant features.

\subsection{Multi-Task Adaptation on the Office Dataset}
While our attribute level adaptation approach is most suitable in the fine-grained setting, we also tested its efficacy on the office dataset~\cite{office_dataset} since there are no other fine-grained adaptation datasets. The office dataset consists of $31$ classes of objects found around the office (such as backpacks, computers, desk lamps and scissors). For each of these objects, images are available from $3$ domains: Amazon, WebCam and DSLR. While this dataset, introduced in 2010, is still the standard adaptation benchmark used today, its size is much smaller than the car dataset used in our experiments. For example, the WebCam domain consists of $785$ images in total (across $31$ classes). 

Since the office dataset does not have attribute level annotations, we use class labels with varying degrees of granularity to evaluate our multi-Task adaptation approach. We annotate each image with the class name of its parent's, grandparent's and great grand parent's node in the WordNet hierarchy~\cite{wordnet}. Thus, each image has $3$ labels consisting of $3$, $7$ and $19$ categories respectively in addition to its class label. We use these additional labels in place of attributes in our multi-task adaptation approach. Our source domain is Amazon and the target is WebCam.

Tab.~\ref{table:office_unsup} and Tab.~\ref{table:office_semi} show our results for the unsupervised and semi-supervised scenarios respectively. Augmenting both baselines with our multi-task adaptation approach improves performance in the unsupervised as well as semi-supervised settings. This shows that our multi-task approach is not simply limited to attributes, and can be used in any scenario with a hierarchy of labels.

Nevertheless, our method's performance gain on the office dataset is much less than on cars. While car attributes are visually informative, WordNet labels might not be. For example, bike and backpack both share the node ``container'' although their visual appearance is very different. Our future work plans to explore additional methods for obtaining visually distinctive attribute labels.

\subsection{Analysis}
Our model results in a significant increase in performance on most fine-grained categories. As shown in Fig.~\ref{fig:acc_gain}, $75\%$ of held-out categories see a gain in accuracy over~\cite{tzeng2015iccv}. Similarly, in the unsupervised setting, our model improves performance on $66\%$ of the target classes. There is no change in accuracy on $14\%$ and $16\%$ of classes while $10\%$ and $18\%$ of classes see a performance drop in the two regimes respectively.

While, as shown in Fig.~\ref{fig:label_dist}, there is a maximum of $500$ labeled target images per fine-grained class, Figs.~\ref{fig:label_dist}(B), (C), and (D) show that some attributes have as many as $12,000$ labeled images. Although we do not use any target images with fine-grained labels for our $85$ held out classes as training data, there are classes in the training data with shared attributes as the test data. Thus, we expect our method to improve accuracy on classes with many attribute labels. Fig.~\ref{fig:results_fig} \textit{top} shows example images for the top $3$ classes with an accuracy gain in the semi-supervised setting on the car dataset. 
These classes have body type minivan, extended cab and SUV: $3$ out of the top $4$ body types with the highest number of labeled target training images. 

\begin{figure}
\includegraphics[width=\linewidth]
{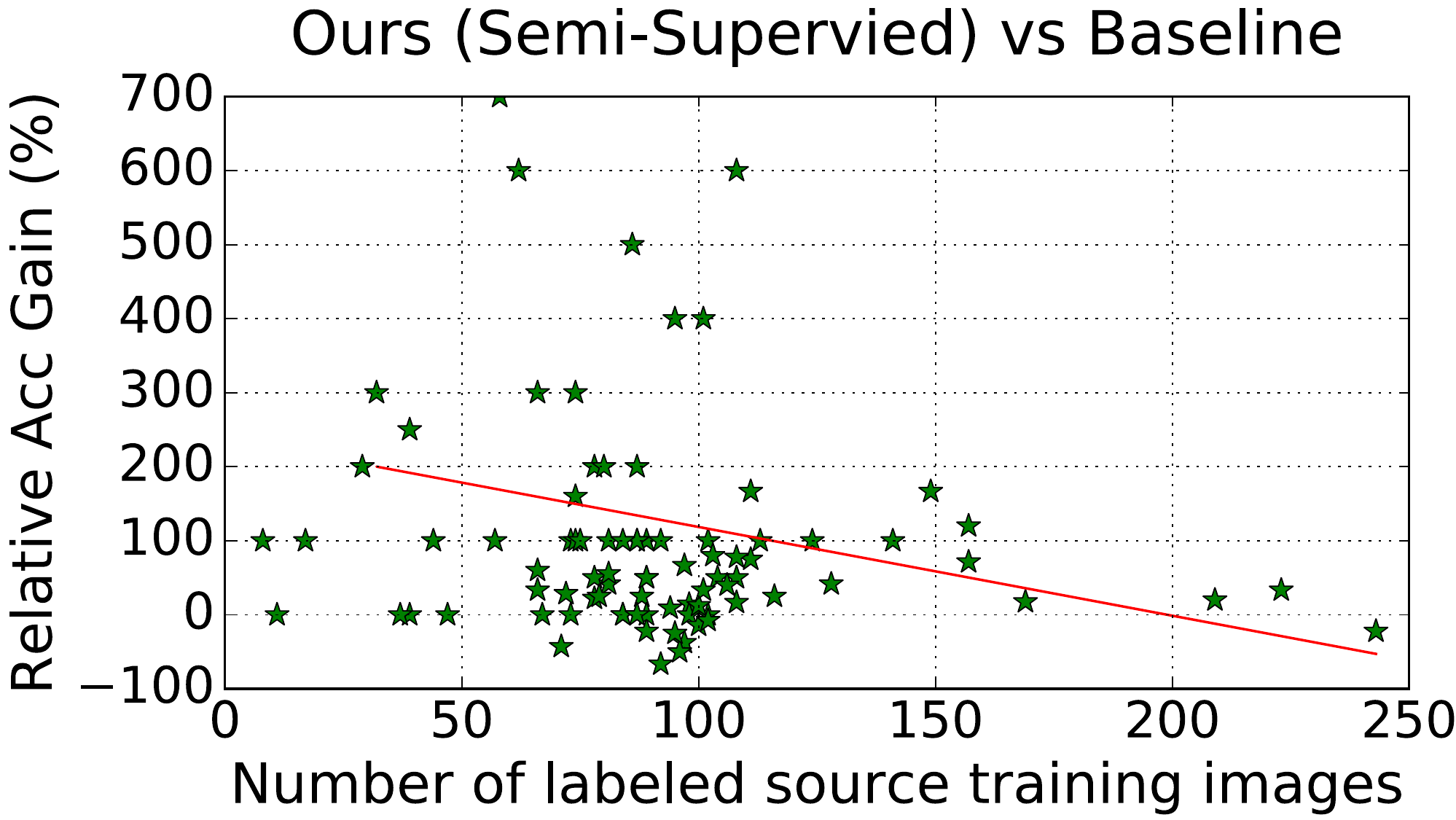}
\caption{The number of labeled images per class vs our relative accuracy gain on the target held-out classes. We see an increase in accuracy gain with decreasing labeled training data.}
\label{fig:acc_label}
\end{figure}

Conversely, the class resulting in the highest accuracy loss with our method is a crew cab: there are only $57$ labeled \textit{GSV} images of crew cabs in our training set. 
Surprisingly, $2$ out of the $3$ classes with the highest accuracy loss are sedans. Although sedans have the 
most number of labeled \textit{GSV} images in our training set (and thus expected to see an accuracy gain), one of these $2$ classes has $243$ source training images. Fig.~\ref{fig:acc_label} plots relative accuracy gain (compared to~\cite{tzeng2015iccv}) vs. the number of labeled source training examples per class. Our approach results in higher accuracy gain on classes with few labeled training data. We measure a correlation of $-0.29$ between the number of labels per class and the accuracy gain. 

\begin{figure}
\includegraphics[width=\linewidth]{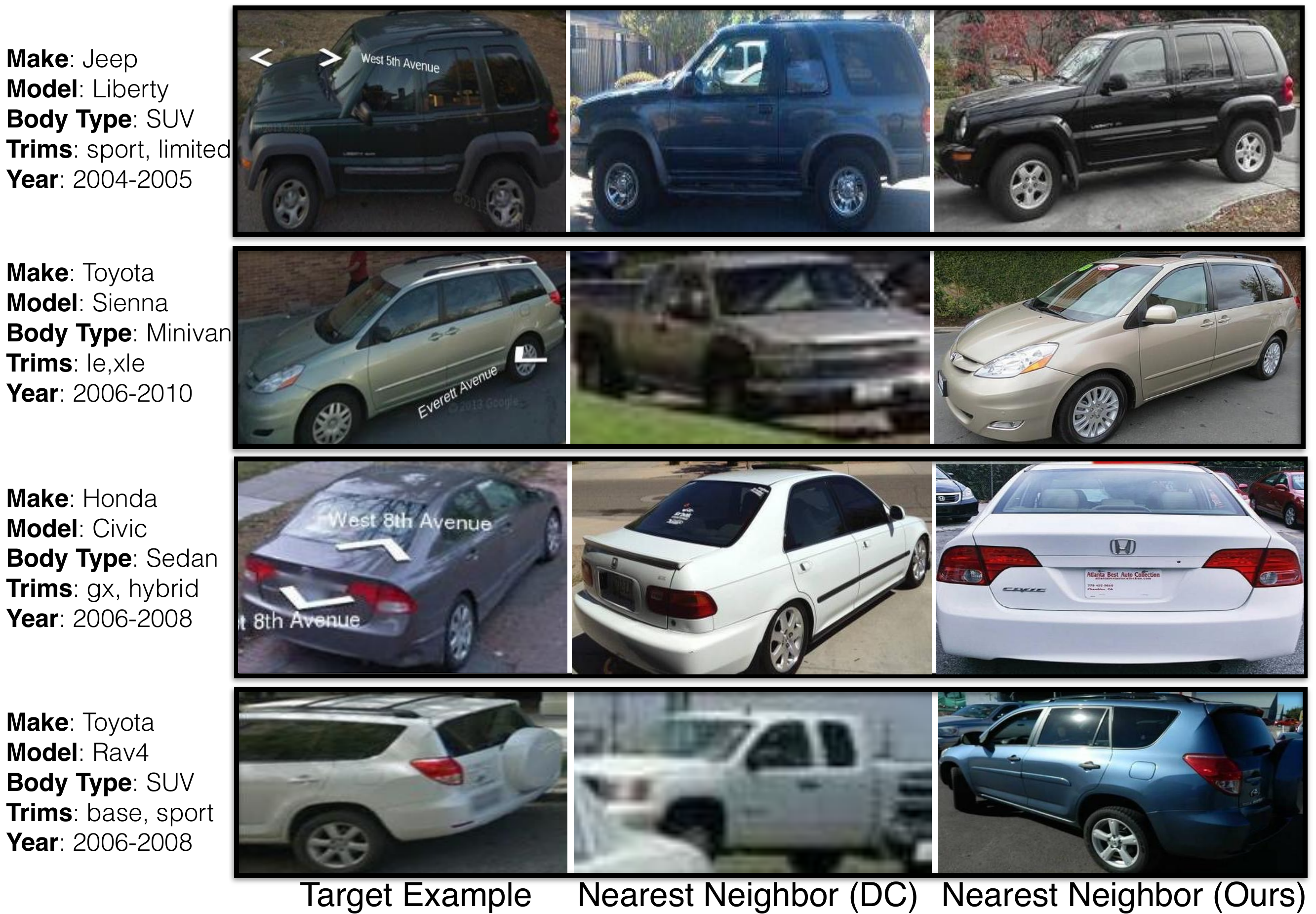}
\caption{Source training images nearest to example target images according to~\cite{tzeng2015iccv} and our multi-task model. Nearest neighbors are computed with $||L_2||$ distance in the feature activation space. First column is the test example, second column shows results of models trained  with~\cite{tzeng2015iccv} to compute the feature activations, and the last column shows results retrieved by our model.}
\label{fig:nearest}
\end{figure}

Finally, Fig.~\ref{fig:nearest} shows example images in the $GSV$ test set and their corresponding nearest neighbors in the training set in the unsupervised setting. For each example image, we compute its feature activations using a baseline model trained with~\cite{tzeng2015iccv}, and our multi-task approach. We retrieve images in the training set whose fc7 activations minimize the $||L_2||$ distance to fc7 activations of the example image. While our attribute based classifier retrieves images in the same class as the target image, the baseline adapted model returns a nearest neighbor in the wrong class. 

\section{Conclusion}
We have presented a multi-task CNN architecture for semi-supervised domain adaptation. Our pipeline leverages the fact that fine-grained classes share attributes which can help transfer knowledge from classes seen in training to those that are not. We evaluated our method on a subset of a large-scale fine-grained dataset consisting of $\sim 1M$ images and $2,657$ car categories. The large number of labeled images from multiple domains makes this dataset ideal for adaptation studies. We also evaluated on the standard office dataset using additional labels from WordNet. In the future, we plan to refine our methodology for incorporating attributes in adaptation, and perform hierarchical adaptation in settings where attribute labels are not available.

\section{Acknowledgments}
We thank Kenji Hata, and Oliver Groth for their valuable feedback. This research is partially supported by an ONR MURI grant, the Stanford DARE fellowship (to T.G.) and by NVIDIA (through donated GPUs). 

{\small
\bibliographystyle{ieee}
\bibliography{egbib}
}

\end{document}